\def\year{2021}\relax
\documentclass[10pt]{article} 
\usepackage{aaai21}  
\usepackage{times}  
\usepackage{helvet} 
\usepackage{courier}  
\usepackage[hyphens]{url}  
\usepackage{graphicx} 
\usepackage{natbib}  
\usepackage{caption} 
\usepackage{times}
\usepackage{latexsym}
\usepackage{mathrsfs}
\usepackage{amssymb}
\usepackage{amsmath}
\usepackage{url}
\usepackage{pifont}
\usepackage{graphicx}
\usepackage{color,xcolor}
\usepackage{braket}
\usepackage{bbold}
\usepackage[utf8]{inputenc}
\usepackage[toc,page]{appendix}
\usepackage{cleveref}
\usepackage{booktabs,graphicx,makecell,siunitx,eqparbox}

\title{Quantum Cognitively Motivated Decision Fusion for Video Sentiment Analysis}
\author{

    Dimitris Gkoumas\textsuperscript{\rm 1},
    Qiuchi Li\textsuperscript{\rm 2}, 
    Shahram Dehdashti \textsuperscript{\rm 3},
    Massimo Melucci\textsuperscript{\rm 2},
    Yijun Yu\textsuperscript{\rm 1},
    Dawei Song\textsuperscript{\rm 1,4}\\

}
\affiliations{

    \textsuperscript{\rm 1} The Open University, Milton Keynes, UK\\
    \textsuperscript{\rm 2} University of Padua, Padua, Italy\\
    \textsuperscript{\rm 3} Queensland University of Technology, Brisbane, Australia\\
    \textsuperscript{\rm 4} Beijing Institute of Technology, Beijing, China\\
    \textsuperscript{\rm 1}\{dimitris.gkoumas, yijun.yu, dawei.song\}@open.ac.uk, \textsuperscript{\rm 2}\{qiuchili, melo\}@dei.unipd.it, \textsuperscript{\rm 3}shahram.dehdashti@qut.edu.au
   
}

\begin{document}

\maketitle

\begin{abstract}
Video sentiment analysis as a decision-making process is inherently complex, involving the fusion of decisions from multiple modalities and the so-caused cognitive biases. Inspired by recent advances in quantum cognition, we show that the sentiment judgment from one modality could be incompatible with the judgment from another, i.e., the order matters and they cannot be jointly measured to produce a final decision. Thus the cognitive process exhibits ``quantum-like'' biases that cannot be captured by classical probability theories. Accordingly, we propose a fundamentally new, quantum cognitively motivated fusion strategy for predicting sentiment judgments. In particular, we formulate utterances as quantum superposition states of positive and negative sentiment judgments, and uni-modal classifiers as mutually incompatible observables, on a complex-valued Hilbert space with positive-operator valued measures. Experiments on two benchmarking datasets illustrate that our model significantly outperforms various existing decision level and a range of state-of-the-art content-level fusion approaches. The results also show that the concept of incompatibility allows effective handling of all combination patterns, including those extreme cases that are wrongly predicted by all uni-modal classifiers.
\end{abstract}

\section{Introduction}

Video sentiment analysis is an emerging interdisciplinary area, bringing together artificial intelligence (AI) and cognitive science. It studies a speaker's sentiment expressed by distinct modalities, i.e., linguistic, visual, and acoustic. At its core, effective modality fusion strategies are in place. Existing neural structures have achieveed the state-of-the-art (SOTA) performance \cite{tsai_etal_2019_multimodal,wang2019words,dumpala2019audio,zadeh2017tensor} by integrating features after being extracted, called \textit{feature-level} fusion. Other approaches simulate logic reasoning and human cognitive biases \cite{morvant2014majority,glodek2011multiple,glodek_2011_senti} by aggregating decisions of uni-modal classifiers into a joint decision, called \textit{decision-level} fusion. Additionally, hybrid fusion approaches benefit from the advantages of both strategies. In this paper, we target at the generally less effective but more flexible decision-level fusion.

Video sentiment analysis is inherently a complex human cognition process. Recent research in cognitive science found that in some cases human decision making could be highly irrational \cite{tversky1983extensional}, and such behaviour does not always obey the classical axioms of probability \cite{kolmogorov1950foundations} and utility theory \cite{morgenstern1949theory}. On the other hand, the mathematical formalism of Quantum Mechanics (QM) has been shown able to address paradoxes of classical probability theory in modelling human cognition \cite{busemeyer2012quantum}. Conceptually, quantum cognition challenges the notion that human's cognitive states underpinning the decisions have pre-defined values and that a measurement merely records them. Instead, the cognitive system is fundamentally uncertain and in an indefinite state. The act of measurement would then create a definite state and change the state of the system. 

We hypothesise that uni-modal sentiment judgments do not happen independently, like a pre-defined value being read out of the internal cognitive state. They are rather constructed at the point of information interaction and thus influenced by the other modalities, which serve as a context for sentiment judgment under the current modality. For example, there might be cases that the order of different decision perspectives, e.g., when someone focuses first on the linguistic and then on the visual view, or vice versa, could lead to controversial sentiment judgements. That is, the measurement from the first perspective provides a context that affects the subsequent one, influencing the probabilities used to compute the utility function of multimodal sentiment decision. In this case, we say that these two decision perspectives are  \textit{incompatible} with each other. Such incompatibility implies that judgements over different modalities cannot be measured jointly, and quantum probability as a generalization of the classical probability theory should hence be in place \cite{uprety2020quantum}. We argue that video sentiment analysis could benefit from the generalised framework of quantum cognition by capturing the cases of incompatibility that cannot be accommodated by classical probabilities.

This paper introduces a novel decision-level fusion strategy inspired by quantum cognition~\cite{fell2019experimental}. The goal is to predict the sentiment of utterances in videos, associated with linguistic, visual, and acoustic streams. We formulate an utterance as a \textit{quantum superposition} state of positive and negative sentiments (i.e., it can be positive and negative at the same time until it is judged under a specific context), and uni-modal classifiers as \textit{mutually incompatible observables}, on a complex-valued Hilbert space $\mathcal{H}$ spanned by distinct uni-modal sentiment bases. We take advantage of incompatibility to influence the uni-modal decisions, when they are under high uncertainty, to finally infer multimodal sentiment judgment. It is important to note that the model produces a generalized form of classical probabilities, allowing for both compatible and incompatible sentiment decisions.

More specifically, we resolve the incompatibility issue through Positive-Operator Valued Measures (POVMs) to approximate the sentiment of uni-modal classifiers simultaneously. In practice, we estimate the complex Hilbert Space and uni-modal observables from training data, and then establish the final multimodal sentiment state of a test utterance from the learned uni-modal observables. To our best knowledge, this is the first quantum cognitively inspired theoretical approach, with practical implementations, that investigates and models the incompatibility of sentiment judgments for video sentiment analysis.

Extensive evaluation on two widely used benchmarking datasets, CMU-MOSI\cite{zadeh2016mosi} and CMU-MOSEI\cite{zadeh2018multimodal}, show that  our strategy significantly outperforms various decision-level fusion baselines and a range of SOTA content-level fusion approaches for video sentiment analysis. The model is also shown able to make correct sentiment judgments even for the cases where all uni-modal classifiers give wrong predictions.
\section{Related Work}

\subsection{Video Sentiment Analysis}
Existing neural structures have achieved SOTA results for utterance-level video sentiment analysis \cite{gkoumas2021makes}. Memory networks incorporate multimodal hidden units of preceding timestamps with the inputs~\cite{wang2019words,liang2018multimodal,zadeh2018memory}. Tensor-based operations have also been exploited to compose~\cite{zadeh2017tensor} or factorize~\cite{liu2018efficient, tensorFactorization} different modalities. Moreover, some approaches introduced fuzzy logic~\cite{chaturvedi2019fuzzy} and encoder-decoder structures in sequence-to-sequence learning, translating a target modality to a source modality~\cite{pham2019found,dumpala2019audio}. Multimodal transformer~\cite{tsai_etal_2019_multimodal} achieved the SOTA in affective analysis tasks. 
While most existing strategies fuse word-level aligned modalities, there are also attempts to address the challenging task of fusing unaligned time-series streams \cite{wang2019words}.

\subsection{Quantum-inspired Representation Learning}
The application of Quantum Theory (QT) in representation learning began after van Rijsbergen’s pioneering work~\cite{van2004geometry} by integrating geometry, probabilities, and logic into a unified theoretical framework. In particular, QT has been exploited beyond its standard domain for various representation learning tasks \cite{uprety2020survey}. Among them, quantum formalism has been utilised for modelling word dependencies through density matrices~\cite{qlm} and formulating the semantic composition of words~\cite{quantum_latent_topics} in Information Retrieval tasks. Recently, researchers deployed quantum probabilistic models to address Natural Language Processing tasks. Preliminary work introduced neural networks to simulate  \textit{quantum compositionality} for question-answering tasks~\cite{ZhangNSWM018}. Later, the simulation of \textit{quantum measurement} postulate and its procedural steps into end-to-end neural networks led to improved performance and better interpretability~\cite{semantic_hilbert,li-etal-2019-cnm}. Furthermore, quantum probabilistic neural models have been exploited to mode interactions across speakers in conversations~\cite{ijcai2019-755}.  Quantum-inspired strategies were also investigated for multimodal representation learning tasks. Wang et al. \cite{wang2010tensor} proposed a tensor-based representation to retrieve image-text documents.
Moreover, there have been studies investigating \textit{quantum interference}~\cite{ZHANG201821} and \textit{non-classical correlations}~\cite{gkoumas2018investigating} to address the decision-level modality fusion. Recently, a quantum-inspired neural framework achieved the SOTA performance for utterance-level video sentiment analysis~\cite{li2020quantum}. Unlike these existing works, in this paper, we propose a quantum cognition inspired theoretical model capturing cognitive biases via the concept of incompatibility, accommodated only by quantum probabilities.

\section{Background of Quantum Cognition}
\label{Sec:Background}
In this section, we introduce the key concepts of quantum cognition \cite{busemeyer2012quantum,fell2019experimental}, which we exploit to construct the proposed model.

\subsection{Hilbert Space}
Quantum cognition exploits an infinite complex-valued vector space, called Hilbert space $\mathcal{H}$, in which the state of a quantum system is represented as a unit-length vector. Different from classical probability, quantum probability events are defined as orthonormal basis states. A projective geometric structure establishes relationships between states vectors and basis states \cite{Halmos87,hughes1989structure}. The same Hilbert space can be represented by different sets of orthonormal basis states, and the same state can be defined over different sets of orthonormal basis states. 

In consistency with QM, we adopt the widely-used \textit{Dirac Notations} for the mathematical formalism of quantum cognition. A complex-valued \textit{unit} vector $\vec{u}$ and its conjugate transpose $\vec{u}^{\ast T}$ are denoted as a \textit{ket} $\ket{u}$ and a \textit{bra} $\bra{u}$, respectively. The inner product of  two  vectors $\vert u \rangle$ and $\vert v \rangle$ is defined by $\braket{u|v}$, while $\ket{u}\bra{u}$ and $\ket{v}\bra{v}$ define operators.

\subsection{Quantum Superposition}
\label{sec:states}
Quantum superposition is one of the fundamental concepts in QM, which describes the uncertainty of a single particle. In the micro world, a particle like a photon can be in multiple mutually exclusive basis states simultaneously with a probability distribution. A general pure state $\ket{\psi}$ is a vector on the unit sphere, represented by
\begin{equation}
    \ket{\psi} = w_1 \ket{e_1} + ...+ w_n \ket{e_n}.
\end{equation}
Where $\{\ket{e_j}\}_{j=1}^n$ are \textit{basis states} forming an orthogonal basis of the Hilbert Space, and the \textit{probability amplitudes} $\{w_j\}_{j=1}^n$ are complex scalars with $\sum_{j=1}^n |w_j|^2 = 1$, and $|\cdot|$ the modulus of a complex number. $\ket{\psi}$ is a \textit{superposition state} when it is not identical to a certain basis state $\ket{e_j}$. 
In particular, in a two-dimensional Hilbert Space $\mathcal{H}_2$ spanned by basis states $\ket{0}$ and $\ket{1}$, a pure state $\ket{\psi}$ is represented as

\begin{equation}
    \label{eq:superpostion}
   \ket{\psi} = \cos\frac{\theta}{2}\ket{0} + e^{i\phi}\sin\frac{\theta}{2}\ket{1}
\end{equation}
while $\theta, \phi \in [0,2\pi]$ and $i$ is the imaginary number satisfying $i^2 = -1$. Eq.~\ref{eq:superpostion} uniquely expresses any pure state on $\mathcal{H}_2$.

\subsection{Measurement}
\label{sec:measurements}
Measurement is another fundamental concept in quantum cognition for calculating quantum probabilities. In QM, Projection-Valued Measure (PVM) removes a system state from uncertainty to a precise event, by projecting a state to its certain corresponding basis state\footnote{For simplicity, we use a nomenclature definition. A strict definition can be found in ~\cite{nielsen_quantum_2011}.}. In the absence of measurement, there is uncertainty in the state in that it takes all possible measurement values simultaneously. After measurement, the state \textit{collapses} onto a certain basis state.  However, PVMs on subsystems of a larger system cannot be described by a PVM acting on the system itself. Positive-Operator Valued Measure (POVM) overcomes this constraint, by associating a positive probability for each measurement outcome, ignoring the post-measurement state \cite{nielsen_quantum_2011}. That is, POVM is a generalization of PVM, providing mixed information of a state for the entire ensemble of subsystems.

Mathematically, a POVM $M$ is a set of \textit{Hermitian  positive semi-definite} operators $\{E_i\}$ on a Hilbert space $\mathcal{H}$ that sum to the identity operator, i.e., $\sum_i E_i=\mathbb{1}$. For a pure state $\ket{\psi}$, we can calculate its density matrix $\rho= \ket{\psi}\bra{\psi}$. The probability with respect to $E_i$ is computed as 

\begin{equation} \label{eq:povm}
    P(i) = Tr(E_i \rho) = \bra{\psi}E_i\ket{\psi}
\end{equation}
and $\sum P(i)=1$. 

In the case of measuring a state on a two-dimensional Hilbert Space $\mathcal{H}_2$ (see Eq. 2), the POVM is associated with the following operators~\cite{busch1986unsharp}:

\begin{eqnarray}
    E_{+} &=& \frac{\eta}{2}\mathbb{I} +  (1-\eta) \ket{1}\bra{1}  \\
    E_{-} &=& \frac{\eta}{2}\mathbb{I} +  (1-\eta) \ket{0}\bra{0} 
\end{eqnarray}

\noindent where $\mathbb{I}$ and $\eta$ stand for the identity matrix and noise parameter respectively. The value $\eta \in [0,1]$ determines the probability that the measurement fails due to the system-apparatus correlation or incompatibility \cite{liang2011specker}. When $\eta = 0$, the measurement apparatus exerts no influence on the measurement, and we have an approximate measurement.  When $\eta = 1$, the output of measurement is completely random.

\subsection{Incompatibility}
The concept of \textit{incompatibility} is applicable to a Hilbert space only.  Each basis state, defining a probability event, has a projector $\Pi$ to evaluate the event. The conjunction of two events is not necessarily commutative \cite{busemeyer2018hilbert}. Suppose $\Pi_A$ and $\Pi_B$ are two sequential measurements for $A$ and $B$ events respectively. In quantum cognition, the joint probability distribution of two events equals the product of the two projectors $\Pi_A$ and $\Pi_B$, corresponding to the basis state $A \cap B$. If $\Pi_{A}\Pi_{B} = \Pi_{B} \Pi_{A}$, then the two events are called \textit{compatible}. However, if $\Pi_A \Pi_B \neq \Pi_B  \Pi_A$, then their product is not a projector, and the two events do not commute, i.e., they are \textit{incompatible}. Incompatibility implies that the two measurements cannot be accessed jointly without disturbing each other. Assuming that measurements are always compatible, classical probability can not capture such disturbance. However, the mathematical formalism of quantum probability allows for both compatible and incompatible measurements \cite{hughes1989structure}. It is a generalization of the classical probability theory. 
\section{Task Formulation}
Due to space limitation, this work targets at the binary video sentiment analysis task. Formally, each utterance $U_i \in \{U_1,...,U_N\}$ is associated with linguistic, visual and acoustic features $U_i = \{X_{i,l}, X_{i,v},X_{i,a}\}$ and a positive or negative sentiment label $y_i \in [-1,1]$. The objective is to establish a function, mapping an utterance $U_i$ to its corresponding sentiment label. 
Note that the proposed fusion strategy is extendable to multiclass classification by adopting a one-vs-all classification strategy.

\section{Quantum Cognition-inspired Fusion Model}
We now introduce the proposed quantum cognition-inspired fusion model for video sentiment analysis. 


\subsection{Sentiment Hilbert Space}
\begin{figure}[t]
\centering
\includegraphics[width=0.4\textwidth]{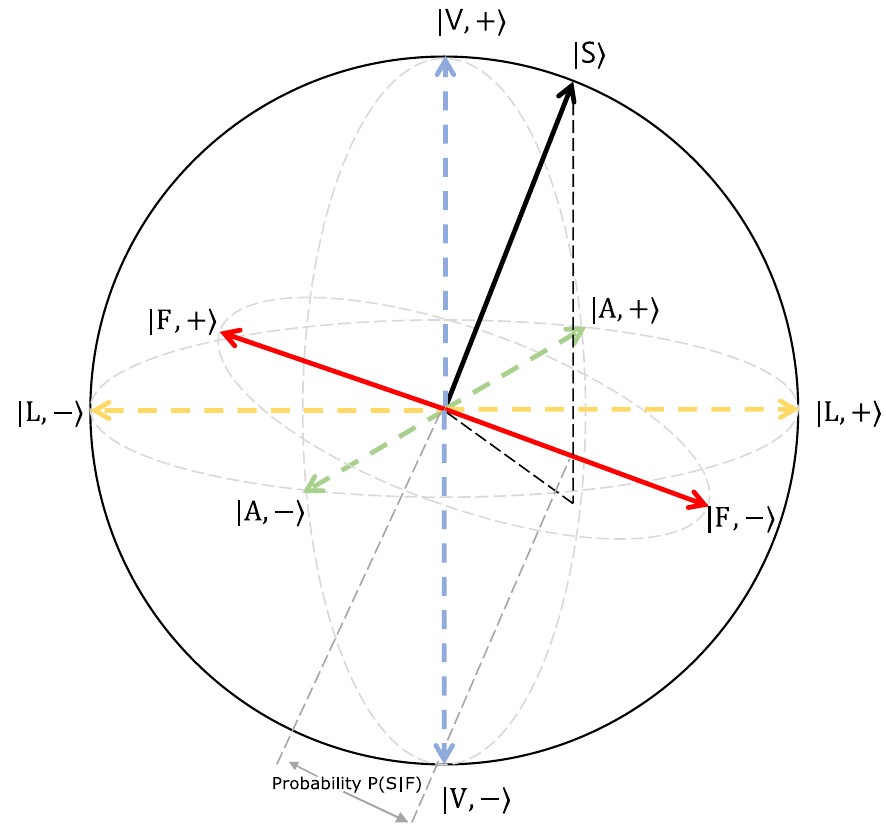}
\caption{The Sentimental Hilbert Space. An utterance is represented as a pure state $\ket{S}$ belonging to the surface of a unit sphere (called the Bloch sphere). The two opposed unit vectors represent positive and sentiment judgments. The associated uni-modal sentiment observables $\hat{L},\hat{V},\hat{A}$ and tri-modal observable $\hat{F}$ are mutually incompatible. Projections of $\ket{S}$ on the corresponding bases indicate probability events.}
\label{fig:bloch_sph}
\end{figure}

The model is defined on a \textit{Sentimental Hilbert Space} $\mathcal{H}_{senti}$, which is a 2-dimensional vector space spanned by basis states $\{\ket{+}, \ket{-}\}$. The basis states $\ket{+}$, $\ket{-}$ correspond to the positive and negative sentiments, respectively. We represent an utterance $U_k$ as a pure state $\ket{S_{U_k}}$ (in short $\ket{S}$)  on $\mathcal{H}_{senti}$. The uni-modal sentiment classifiers (denoted as $L, V, A$ respectively) are formulated as mutually incompatible observables (see Figure \ref{fig:bloch_sph}). The utterance can be represented under different sets of basis states, i.e., uni-modal ($L, V, A$) and multimodal ($F$) basis states in Figure \ref{fig:bloch_sph}. The observables are not orthogonal with each other, since the modalities are not independent, but highly correlated.

\subsection{Utterance Representation}
An utterance is represented as a pure state $\ket{S}$  of positive and negative sentiments on a 2-dimensional Hilbert space $\mathcal{H}_{senti}$ (see Figure \ref{fig:bloch_sph}):
\begin{equation}\label{eq:sen_state}
    \ket{S} = \cos\frac{\theta_S}{2}\ket{+} + e^{i\phi_S}\sin\frac{\theta_S}{2}\ket{-}
\end{equation}
\noindent where $\theta_S, \phi_S \in [0,2\pi]$. According to the Born's rule \cite{born1926quantenmechanik}, the probability of an utterance being positive and negative is $P{(+)} =| \braket{S|+}|^2 = \cos^2\frac{\theta_S}{2}$ and $P{(-)} = |\braket{S|-}|^2 = \sin^2\frac{\theta_S}{2}$, with $\cos^2\frac{\theta_S}{2} + \sin^2\frac{\theta_S}{2} = 1$. As to be shown in more detail in the next Section, the \textit{relative phase} $\phi_S$ plays a crucial role in capturing correlations between incompatible observables and giving rise to results that are fundamentally different from the classical case.

\subsection{Sentiment Decisions}
We formulate uni-modal sentiment decisions as mutually incompatible observables on $\mathcal{H}_{senti}$, namely $\hat{L}$, $\hat{V}$, and $\hat{A}$ for linguistic, visual and acoustic modalities respectively (see Figure \ref{fig:bloch_sph}). For the binary sentiment analysis task, each observable is associated with two eigenstates and two eigenvalues, with common eigenvalues of $1$ and $-1$ for positive and negative sentiments. In that case, incompatibility falls under different sets of eigenstates $\{\ket{M,+},\ket{M,-}\}$ defining a uni-modal basis, where modality $M \in \{L,V,A\}$. Following Eq.~\ref{eq:sen_state},  we express the eigenstates as 

\begin{eqnarray}
    \ket{M,+} &=& \cos\frac{\theta_M}{2}\ket{+} + e^{i\phi_M}\sin\frac{\theta_M}{2}\ket{-} \label{m_plus} \\
    \ket{M,-} &=& \sin\frac{\theta_M}{2}\ket{+} - e^{i\phi_M}\cos\frac{\theta_M}{2}\ket{-}
\end{eqnarray}

\noindent with $\theta_M, \phi_M \in [0,2\pi]$. The eigenstates form an orthonormal basis, with $\braket{M,+|M,+} = \braket{M,-|M,-} = 1$ and $\braket{M,+|M,-} = \braket{M,-|M,+} =  0$.  

In QT, a general observable $\hat{O}$ can be decomposed to its eigenstates $\{\ket{\lambda_i}\}$ of the orthonormal basis as $\hat{O} = \lambda_i \ket{\lambda_i} \bra{\lambda_i}$, where eigenvalues $\{\lambda_i\}$ are possible values that a state can take for the corresponding events after measurement. Thus, the uni-modal observables have as follows:  
\begin{eqnarray}
    \hat{M} &=& (+1)\ket{M,+}\bra{M,+} +(-1)\ket{M,-}\bra{M,-}  
\end{eqnarray}
where $\hat{M} \in \{\hat{L}, \hat{V}, \hat{A}\}$. 
Similarly, the observable for the final sentiment decision $\hat{F}$ is 
\begin{equation} \label{eq.final_state}
     \hat{F}=(+1)\ket{+}\bra{+}+ (-1)\ket{-}\bra{-}
\end{equation}
which spans the $\mathcal{H}_{senti}$ and is incompatible with all uni-modal observables. 

Following the projective geometric structure, the measurement probability on an eigenstate equals the projection of the system state onto it, i.e., the squared inner product of the vectors: $|\braket{S|M, +}|^2$ for uni-modal positive sentiment and $|\braket{S|+}|^2$ for final (multimodal) positive sentiment. The measurement probabilities under $\hat{L}$ stand for the utterance's sentiment under linguistic modality, as so forth for the other modalities. Finally, its multimodal sentiment is determined by the observable $\hat{F}$. In Figure \ref{fig:bloch_sph}, the sentiment judgment is positive in terms of uni-modal observables (projections are visualized as shadows in Figure \ref{fig:bloch_sph}), yet is negative in terms of the multimodal observable due to incompatibility.  

\section{Model Operationalization}
This section presents a methodology that operationalizes the proposed fusion model. Traditionally, in the physical sciences, the study of mathematical problems involves modelling methods leveraging a combination of approximation techniques. In this work, we exploit statistical information from the data to learn the sentimental Hilbert Space described in the previous section, so as to leverage the incompatible observables to determine the sentiment polarity of utterances.  We propose a pipeline consisting of three steps: (1) we first estimate the generic uni-modal observables $\hat{M}$ from the training data; (2) then we construct the sentiment state for each test utterance $\ket{S_T}$ from the learned uni-modal observables and uni-modal sentiment prediction results; (3) finally, we judge the sentiment with the multimodal observable $\hat{F}$. In the remaining part of the section, we elaborate the methodology for each step.

\subsection{Observable Estimation}
The uni-modal observables are constructed from the overall statistics of the training data. These values are mapped to their quantum expressions to estimate the parameters of the uni-modal observables.  In particular, the uni-modal observables and pure state should submit to the following properties: I) the pure state should conform to the statistics of the dataset, II) the uni-modal sentiment measurement results should be consistent with the ratio of positive and negative samples in the training subsets, and  III) quantum correlations between observables should be aligned to classical correlations of the per-sample prediction results, derived from the training data. 

To facilitate the construction of uni-modal observables, we introduce a pure state as follows:
\begin{eqnarray} \label{eq:general_state}
\ket{G} = \cos{\frac{\theta_{G}}{2}}\ket{+}+e^{i\phi_{G}} \sin{\frac{\theta_{G}}{2}}\ket{-}
\end{eqnarray}
which describes the extent to which the dataset is unbalanced for positive and negative labels. By Born's rule \cite{born1926quantenmechanik}, the probability of positive judgment is:
\begin{eqnarray}  \label{eq:mm_mea_1}
  P(+) = |\braket{+|G}|^2 \approx \frac{\# pos}{N}
\end{eqnarray}
where $\#pos$ is the number of true positive utterances in the training set and $N$ the size of the training set. Eq.\ref{eq:mm_mea_1} implies
\begin{eqnarray} \label{eq:mm_mea_2}
\cos^2{\frac{\theta_{G}}{2}} \approx \frac{\# pos}{N}
\end{eqnarray}
since the quantum probability equals the squared amplitude of a state (see Section Background).

According to the second property, the probability of a positive sentiment judgment for each modality is given by
\begin{eqnarray}
    \label{eq:m_g_plus}
     P_M(+) = |\braket{M,+|G}|^2  \approx \frac{\#M_{pos}}{N}
\end{eqnarray}
where $\#L_{pos}$, $\#V_{pos}$ and $\#A_{pos}$ equals the number of true positive utterances for each modality in the training set. Combining Eq.\ref{m_plus}, Eq.\ref{eq:general_state} and Eq.\ref{eq:m_g_plus}, the probability of the positive sentiment judgment for each modality is 
\begin{equation} 
\label{eq:um_mea}
\begin{aligned}
 &\cos^{2}\frac{\theta_{M}}{2}\cos^{2}\frac{\theta_{G}}{2}+\sin^{2}\frac{\theta_{M}}{2}\sin^{2}\frac{\theta_{G}}{2}  \\
  &+ \frac{1}{2}\sin \theta_{M}\sin\theta_{G}\cos (\phi_{M}-\phi_{G}) \approx \frac{\#m_{pos}}{N}
\end{aligned} 
 \end{equation}

Finally, we look into the correlations between pairs of uni-modal observables, where the relative phases play an key role. From a QM point of view, the correlation of observables for two modalities $M_1, M_2$ is given by $(|\braket{M_1,+|M_2,+}|^2 + |\braket{M_1,-|M_2,-}|^2 - |\braket{M_1,+|M_2,-}|^2 - |\braket{M_1,-|M_2,+}|^2)$. It should in principle be aligned to the classical correlations derived from the data. Hence we have
\begin{equation} 
\label{eq:corr}
\begin{aligned}
&\frac{1}{2}\big(|\braket{M_1,+|M_2,+}|^2 + |\braket{M_1,-|M_2,-}|^2  \\ &-  |\braket{M_1,+|M_2,-}|^2  - |\braket{M_1,-|M_2,+}|^2\big)
     \\ &\approx corr(M_1,M_2)
\end{aligned}    
\end{equation}
where $M_1 \neq M_2 \in \{L,V,A\}$ and  $corr(M_1,M_2)$ is a classical correlation of the per-sample prediction results based on modalities $M_1$ and $M_2$, which is computed from the training data. When $M_1$ and $M_2$ give exactly same predictions, the correlation $corr(M_1,M_2) = 1$. Accordingly, $|\braket{M_1,+|M_2,+}| =  |\braket{M_1,-|M_2,-}|^2 = 1$ and $|\braket{M_1,+|M_2,-}|^2 = |\braket{M_1,-|M_2,+}|^2 = 0$, so the value 1 is also produced from the quantum side.  Similarly, a value of -1 is obtained for both sides when the two modalities give totally opposite predictions, indicating the maximum negative correlation. Hence, Eq.\ref{eq:corr} gives three equations for distinct pairs of modalities. For example, for linguistic-visual correlation, Eq.\ref{eq:corr} results in
\begin{equation} 
\label{eq:corr_ex}
\begin{aligned}
\cos\theta_{L}\cos\theta_{V} + \sin\theta_{L}\sin\theta_{V}\cos(\phi_{L}-\phi_{V}) \\ \approx  corr(L,V)
\end{aligned}    
\end{equation}
as so forth for the $\{L,A\}, \{A,V\}$ modality pairs. 

To wrap up, taking into account the number of positive sentiments in the training set and correlations across different pairs of modalities, we get seven equations from Eq.~\ref{eq:mm_mea_2}, Eq.~\ref{eq:um_mea} and Eq.~\ref{eq:corr}, and eight unknown variables $\{\theta_G,\theta_L,\theta_V,\theta_A,\phi_G,\phi_L,\phi_V,\phi_A\}$. As all equations rely only on the differences between the relative phases rather than their absolute values, we can safely set $\phi_g = 0$ and devise a unique solution of $\{\theta_l,\theta_v,\theta_a,\phi_g,\phi_l,\phi_v,\phi_a\}$. Through solving the system of equations, we calculate the parameters for each uni-modal observable $\hat{L}$, $\hat{V}$, and $\hat{A}$.

\subsection{Utterance State Estimation}
After estimating uni-modal observables as described above, we need to estimate the state for each test utterance, which is denoted as
\begin{eqnarray} \label{eq:test_set}
    \ket{S_T} = \cos\frac{\theta_{T}}{2}\ket{+} + e^{i\phi_{T}}\sin\frac{\theta_{T}}{2}\ket{-}
\end{eqnarray}
The uni-modal predictions can be exploited to estimate the values of $\theta_T, \phi_T$. However, since the observables $\hat{L}$, $\hat{V}$, and $\hat{A}$ are mutually incompatible, the measurements results cannot be accessed simultaneously. To that end, we utilize POVMs to get the results of all incompatible measurements simultaneously \cite{uola2016adaptive}. In particular,  we construct sample-specific POVMs for each uni-modal measurement, applying unsharp (weak) projections \cite{busch1986unsharp} without disturbing the observables. 
Formally, the operators are constructed as follows:  
\begin{eqnarray}
    E_{\pm}^{M}=\frac{\eta_T}{2}\mathbb{I}+ (1-\eta_T) \ket{M,\pm}\bra{M,\pm}
\end{eqnarray}

\noindent where $\eta_T \in [0,1]$ is specific to sample $T$, since each utterance interacts with the apparatus in a different manner. We apply  uni-modal POVMs on the test utterance to measure its sentiment in terms of each modality, that is,
\begin{eqnarray}
\label{eq_uni_povs}
    \bra{S_T}E_{+}^M\ket{S_T} \approx P_{T,M}(+)
\end{eqnarray}
where $P_{T,M}(+)$ are uni-modal probabilities for the positive sentiment judgment. Eq.\ref{eq_uni_povs} gives a system with three equations, each equation for a distinct modality, and three unknown variables $\{\theta_T, \phi_T, \eta_T\}$. Solving the system allows us to construct the state $\ket{S_T}$. 

\subsection{Multimodal Sentiment Measurement}
The sentiment of a test utterance $\ket{S_T}$ is measured by Eq. \ref{eq.final_state}. The results are $P_T(+) = \cos^2\frac{\theta_{T}}{2}$ and $P_T(-) = \sin^2\frac{\theta_{T}}{2}$. The sentiment of $S_T$ is considered as positive if $\cos^2\frac{\theta_{T}}{2} > 0.5$ and negative otherwise.

\section{Experiments}
We evaluated the proposed model on two benchmarking datasets, namely, CMU Multimodal Opinion-level Sentiment Intensity (CMU-MOSI)~\cite{zadeh2016mosi} and CMU Multimodal Opinion Sentiment and Emotion Intensity (CMU-MOSEI)~\cite{zadeh2018multimodal}. Each sample is labelled with a 7-level ratio score. In this work, we adopted binary accuracy (i.e., $Acc_{2}:$ positive sentiment if the human annotation score $\geq 0$, and negative sentiment if the score $ < 0$),  and $F1$ score. For both datasets, we used the CMU-Multi-modal Data SDK\footnote{https://github.com/A2Zadeh/CMU-MultimodalSDK} for feature extraction.

\subsection{Baselines}
We compared with robust approaches on both decision-level and feature-level modality fusion approaches.
\subsubsection{Decision-level:} We first trained neural uni-modal classifiers. In particular, we used Bi-GRU layers~\cite{cho2014learning} with forward and backward state concatenation, followed by fully connected layers. The outputs gave linguistic, visual, and acoustic embedding $\{L, V, A\} \in \mathbf{R}^d$, where $d$ was the number of neurons in dense layers. Then, self-attentions were computed for each uni-modal dense representation by calculating the scaled dot-product~\cite{vaswani2017attention}. Finally, each attentive uni-modal representation was fed into two fully connected layers, followed by a softmax layer to obtain sentiment judgments. The uni-modal results were then fed into the multimodal meta-fusion approaches. We compared with a range of baseline fusion approaches:

\begin{itemize} 
    \item \textbf{Voting} was used to aggregate the outputs of the uni-modal classifiers. In particular, we applied a) \textit{Hard Voting}, via majority voting, b) \textit{Weighted Majority Voting}, by assigning weights to each uni-modal classifier and taking their average, and c) \textit{Soft Voting}, by averaging the predicted probabilities, to infer multi-modal sentiment judgments.
    \item \textbf{Single models} exploited supervised machine learning algorithms as meta fusion approaches of the uni-modal classifiers. For both tasks, we chose the most effective models, namely, a) \textit{Logistic Regression}, b) \textit{Support Vector Machine (SVM)}, and c) \textit{Gaussian Naive Bayes (GaussianNB)}, from a pool of supervised learning algorithms.
    \item \textbf{Ensemble methods} combined learning algorithms, selecting the optimum combination from a pool of models. We explored stacking, backing, and boosting strategies \cite{ponti2011combining}; a) for stacking, single models were stacked together and the hard voting method computed predictions, b) for bagging, a number of estimators were aggregated by majority voting, and c) for boosting, we applied AdaBoost classifier~\cite{freund1997decision}
    \item \textbf{A Deep Fusion} approach combined the confidence scores of uni-modal classifiers along with the complementary scores as inputs to a deep neural network, followed by a sigmoid layer, which made the final prediction~\cite{nojavanasghari2016deep}.
\end{itemize}

\subsubsection{Content-level:} We also compared the model with a range of SOTA content-level fusion approaches.
\begin{itemize} 
    \item For \textbf{ SOTA,} we replicated a) \textit{MulT}~\cite{tsai_etal_2019_multimodal}, consisting of pairwise crossmodal transformers, the outputs of which are concatenated to build the multimodal embedding utterance, b) \textit{RAVEN}~\cite{wang2019words}, an RNN based model with an attention gating mechanism to model crossmodal interactions, and c) \textit{TFN}~\cite{zadeh2017tensor}, a tensor-based neural network that a multi-dimensional tensor captures uni-modal, bi-modal, and tri-modal interactions across distinct modalities.
    \item \textbf{QMF}~\cite{li2020quantum} is a complex-valued neural netowork, which represents utterances as superposed states, and incorporates modalities through the tensor operator.
\end{itemize}



\subsection{Experiment Settings}
We conducted the experiments on the same uni-modal classifiers trained for the decision-level baseline approaches. We estimated the uni-modal observables from training plus validation sets, and then we used the learnt observables for predicting the utterance sentiment on the test set. We used Pearson correlation for modelling classical correlations. In case the equation systems did not have solutions, the MATLAB $fsolve$ function was used to generate a numerical solution. In particular, we randomly initialized the parameters $\{\theta_G,\theta_L,\theta_V,\theta_A,\phi_L,\phi_V,\phi_A\} \in [0,2\pi]$ for uni-modal observable estimation, and $\{\theta_T, \phi_T\} \in [0,2\pi], \eta_T \in [0,1]$ for utterance state estimation. The random initialization was repeated for $200$ times to obtain the optimum solutions by calculating the minimum sum of squared loss.

\subsection{Comparative Analysis of Results}
For both datasets, we present the results of the proposed model in comparison with various baseline decision-level fusion strategies in Table \ref{decision_level}. For CMU-MOSEI, all approaches attained an improved performance as compared to the performance of CMU-MOSI task. We suspect this is because CMU-MOSEI is a much larger dataset. Overall, Weighted Voting was the best-performing approach among the voting-based aggregations, Logistic Regression among the supervised learning algorithms, and Stacking among the ensemble learning methods. For both tasks, Stacking and Bagging were the most effective baseline decision-level fusion strategies. For CMU-MOSI, the proposed model attained an increased accuracy of 84.6$\%$ as compared to 78.4$\%$ of Stacking, which was a significant improvement of 6.2$\%$ ($p-value < 0.05$). For CMU-MOSEI, the model reached an increased accuracy of 84.9$\%$ as compared to 82.2$\%$ of Stacking, i.e., a significant improvement of $2.7\%$ ($p-value < 0.05$).  

\begin{table}[ht]
\fontsize{9}{9}\selectfont
\centering 
  \begin{tabular}{lSSSS}
    \toprule
      \multicolumn{1}{c}{} &
      \multicolumn{2}{c}{CMU-MOSI} &
      \multicolumn{2}{c}{CMU-MOSEI} \\
      \cmidrule(r){2-3} \cmidrule(r){4-5}
       Approach & \textbf{$Acc_2$}  & \textbf{$F1$}   & \textbf{$Acc_2$}  & \textbf{$F1$} \\
      \hline
      Hard Voting & {67.5} & {65.4} & {71.5} & {83.3}   \\
      Weighted Voting & {74.6} & {71.6} & {81.3} & {87.8}   \\
      Soft Voting & {75.2} & {71.9} &{77.5} & {86.2}  \\
      \hline
      SVM & {77.4} & {72.9} & {81.7} & {87.9}   \\
      Logistic Regression & {78.0} & {73.8} & {81.9} & {88.0}   \\
      GaussianNB & {76.7} & {71.6} & {80.9} & {86.8}  \\
       \hline
       Stacking & {78.4} & {75.1} & {82.2} & {88.1}   \\
       Bagging & {78.1} & {73.6} & {82.0} & {88.0}   \\
       Boosting & {77.7} & {74.0} & {81.7} & {87.7}\\
       
        \hline
        Deep Fusion  & {77.8} & {77.7} & {81.9} & {81.3}   \\
       \hline
        Proposed Model & \textbf{84.6}  & \textbf{84.5}   & \textbf{84.9}  & \textbf{91.1}    \\
    \bottomrule
  \end{tabular}
  \caption{Effectiveness of decision-level fusion approaches. Best results are highlighted in boldface.} 
  \label{decision_level}
\end{table}

Table \ref{feature_level} presents the comparison results against various SOTA content-level fusion approaches. For CMU-MOSI, TFN \cite{zadeh2017tensor} was the most effective among the content-level fusion baselines. The proposed model attained an improvement in accuracy by 3.4$\%$ (see Table  \ref{feature_level}). For CMU-MOSEI, RAVEN \cite{wang2019words} achieved the highest accuracy among the baselines. The proposed model yielded an increased accuracy of 84.9$\%$ as compared to 80.2$\%$ of RAVEN, i.e., a 4.7$\%$ improvement. 

We noticed that the decision-level fusion strategies achieved better performance than the content-level neural approaches on CMU-MOSEI. This implies that discriminative learning approaches can benefit from large datasets, whereas neural approaches lead to overfitting. We also observed that the proposed model achieved a similar performance on CMU-MOSI and CMU-MOSEI, even though  CMU-MOSI is a relatively balanced dataset. That is, our model can cope with both skewed and balanced datasets.

\begin{table}[htbp]
\fontsize{9}{9}\selectfont
\centering 
  \begin{tabular}{lSSSS}
    \toprule
      \multicolumn{1}{c}{} &
      \multicolumn{2}{c}{CMU-MOSI} &
      \multicolumn{2}{c}{CMU-MOSEI} \\
      \cmidrule(r){2-3} \cmidrule(r){4-5}
       Approach & \textbf{$Acc_2$}  & \textbf{$F1$}   & \textbf{$Acc_2$}  & \textbf{$F1$} \\
       \hline
       MulT \cite{tsai_etal_2019_multimodal} & {80.2} & {79.5} & {80.0} & {79.8}   \\
       RAVEN \cite{wang2019words} & {78.6} & {78.6}   & {80.2} & {79.9}   \\
       TFN \cite{zadeh2017tensor} & {81.2} & {80.8} & {77.8} & {77.8}   \\
       \hline
       QMF \cite{li2020quantum} & {80.7} & {79.7} & {79.7} & {79.6}   \\
       \hline
       Proposed Model & \textbf{84.6}  & \textbf{84.5}   & \textbf{84.9} & \textbf{91.1}   \\
    \bottomrule
  \end{tabular}
  \caption{Effectiveness of content-level fusion approaches.} 
  \label{feature_level}
\end{table}

\subsubsection{Ablation Tests}Table \ref{unimodal_sentiment} shows the results of an ablation study. The first three rows list the performance of uni-modal classifiers when no crossmodal interactions were modelled. The linguistic modality was the most predictive due to the use of word embedding trained on large corpora. For CMU-MOSEI, the linguistic classifier even outperformed all the content-level and voting-based fusion approaches.

\begin{table}[ht]
\fontsize{9}{9}\selectfont
\centering 
  \begin{tabular}{lSSSS}
    \toprule
      \multicolumn{1}{c}{} &
      \multicolumn{2}{c}{CMU-MOSI} &
      \multicolumn{2}{c}{CMU-MOSEI} \\
      \cmidrule(r){2-3} \cmidrule(r){4-5}
       Approach & \textbf{$Acc_2$}  & \textbf{$F1$}   & \textbf{$Acc_2$}  & \textbf{$F1$} \\
      \hline
      Linguistic Only  & {77.1} & {72.3} & {81.5} & {87.8}   \\
      Visual Only  & {54.7} & {48.4} & {71.1} & {83.0}   \\
      Acoustic Only  & {56.1} & {60.0} & {71.2} & {83.1}  \\
      \hline
        Proposed Model & \textbf{84.6}  & \textbf{84.5}  & \textbf{84.9} & \textbf{91.1}     \\
    \bottomrule
  \end{tabular}
  \caption{ Comparison with uni-modal approaches.} 
  \label{unimodal_sentiment}
\end{table}

As a second set of ablation experiments, we tested the proposed model when only bimodal dynamics were present. We present the result in Table \ref{variants}, which shows the linguistic and acoustic dynamics were the most informative. However, trimodal dynamics outperformed all possible bimodal combinations, yielding an improvement of accuracy by $5.0\%$ for CMU-MOSI, and $2.2\%$ for CMU-MOSEI.

\begin{table}[ht]
\fontsize{9}{9}\selectfont
\centering 
  \begin{tabular}{lSSSS}
    \toprule
      \multicolumn{1}{c}{} &
      \multicolumn{2}{c}{CMU-MOSI} &
      \multicolumn{2}{c}{CMU-MOSEI} \\
      \cmidrule(r){2-3} \cmidrule(r){4-5}
       Approach & \textbf{$Acc_2$}  & \textbf{$F1$}   & \textbf{$Acc_2$}  & \textbf{$F1$} \\
      \hline
      Model$_{\{L, V\}}$  & {78.2} & {74.3} & {82.1} & {88.4}   \\
      Model$_{\{L, A\}}$  & {79.6} & {75.1} & {82.7} & {89.2}   \\
      Model$_{\{V, Ac\}}$  & {55.1} & {55.2} & {70.8} & {82.7}  \\
      \hline
        Proposed Model & \textbf{84.6}  & \textbf{84.5}   & \textbf{84.9}  & \textbf{91.1}    \\
    \bottomrule
  \end{tabular}
  \caption{ Comparison of the model with its variants.} 
  \label{variants}
\end{table}

\subsubsection{Effect of Incompatibility}
We conducted a further analysis to investigate the effectiveness of incompatibility. We first identified all the cases that were correctly predicted by one out of the eleven decision-level fusion approaches. In total, there were 33 such cases on CMU-MOSI and 1547 ones on CMU-MOSEI test sets. The proposed model gave correct predictions for 31 cases out of 33 on CMU-MOSI and all the 1547 cases on CMU-MOSEI. Furthermore, we analyzed the cases that all uni-modal classifiers gave wrong sentiment judgments, but the proposed model successfully fused them and gave correct predictions. There were 39 such utterances out of 686 on the CMU-MOSI and 633 utterances out of 4643 on the CMU-MOSEI subsets. 

\subsubsection{Case Study}
\begin{figure}[t]
\centering
\includegraphics[width=0.4\textwidth]{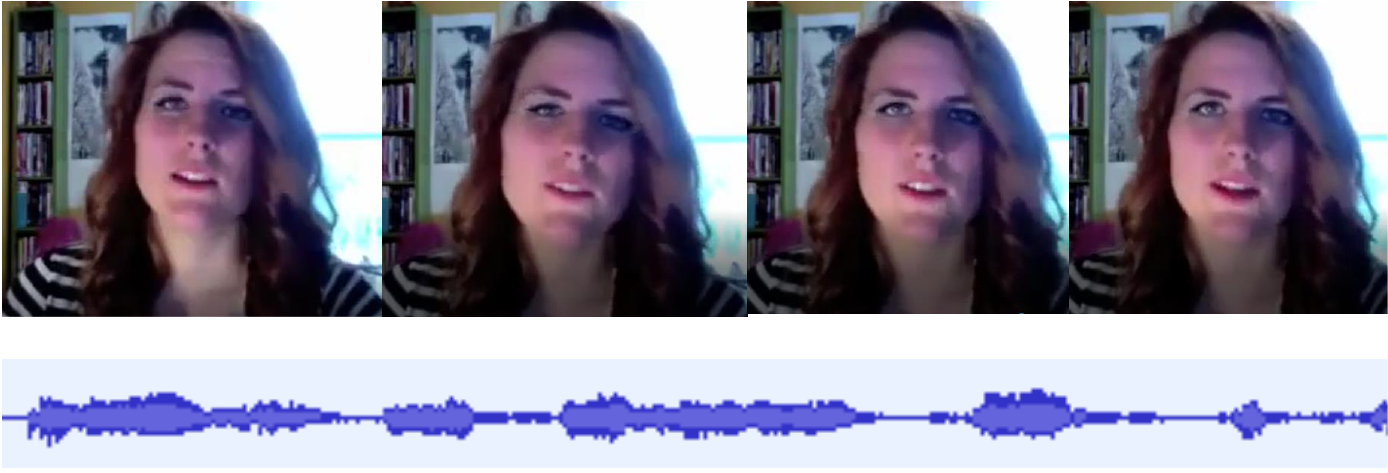}
\caption{Visual-acoustic content of an incompatible case.}
\label{fig:casestudy}
\end{figure}

We illustrate the visual-acoustic content of an incompatible case of the utterance \textit{``I mean even if you don’t have kinds"} in Figure \ref{fig:casestudy}. The linguistic state by itself is in an indefinite state,  which results in a superposition of sentiment judgments. Similarly, the visual-acoustic content is under uncertainty since the content is neutral. Indeed, all uni-modal classifiers predicted a negative sentiment judgment, inferring a probability less than 0.5, yet very close to the decision boundary of 0.5. This superposition of uni-modal beliefs, i.e., positive and negative sentiment at the same time until they are judged under a specific context,  results in the occurrence of incompatibility. Under the high levels of uncertainty, incompatibility influences uni-modal judgments and successfully predicts a positive multimodal sentiment judgment.  This phenomenon is the core of the model and the reason it achieves such high performance.


\section{Conclusions}
We have introduced an effective fusion strategy inspired by quantum cognition.  We formulated utterances as states and uni-modal decisions as mutually incompatible observables in a complex-valued sentimental Hilbert space. The incompatibility captures cognitive biases in the decision fusion process that are otherwise not possible with classical probability. The proposed model has been shown able to handle all combination patterns, including the cases where all uni-modal classifiers gave wrong sentiment judgments. Therefore, the proposed approach achieved an improved performance over SOTA content-level and decision-level modality fusion approaches. In the future, we will investigate the model on conversational video emotion recognition tasks.



\newpage
\section*{Acknowledgements}
This study is supported by the Quantum Information Access and Retrieval Theory (QUARTZ) project, which has received funding from the European Union's Horizon 2020 research and innovation programme under the Marie Sk\l{}odowska-Curie grant agreement No. 721321, and Natural Science Foundation of China (grant No.: U1636203). We thank Christina Amalia Lioma that gave us constructive comments to improve the paper.

\bibliography{main}


\end{document}